\pdfoutput=1
\documentclass[10pt,twocolumn,letterpaper]{article}

\usepackage{iccv}
\usepackage{times}
\usepackage{epsfig}
\usepackage{graphicx}
\usepackage{amsmath}
\usepackage{amssymb}
\def\bs{\boldsymbol}
\usepackage{algorithm}  
\usepackage{algorithmic}

\usepackage{bbm}
\usepackage{booktabs} % for professional tables
\usepackage{setspace}
\usepackage{adjustbox}
\usepackage{multirow}
\def\arrvline{\hfil\kern\arraycolsep\vline\kern-\arraycolsep\hfilneg}

\usepackage{csquotes}

\include{figure}
\usepackage[breaklinks=true,bookmarks=false]{hyperref}

\iccvfinalcopy % *** Uncomment this line for the final submission

\def\iccvPaperID{9778} % *** Enter the ICCV Paper ID here

\ificcvfinal\fi

\begin{document}

%%%%%%%%% TITLE
\title{Improving Text-to-Image Synthesis Using Contrastive Learning}

\author{Hui Ye$^1$, Xiulong Yang$^1$, Martin Tak\'{a}{\v c}$^{2}$, Rajshekhar Sunderraman$^1$, Shihao Ji$^1$ \\
$^1$Department of Computer Science, Georgia State University, Atlanta, GA, USA\\
$^2$Mohamed bin Zayed
University of Artificial Intelligence (MBZUAI), Masdar City,
Abu Dhabi,
UAE\\
% $^2$Computer Science, Georgia State University\\
$^1${\tt\small \{hye2, xyang22\}@student.gsu.edu, $^1$\{rsunderraman, sji\}@gsu.edu}
\\
$^2${\tt\small takac.MT@gmail.com}
}

\maketitle
\ificcvfinal\thispagestyle{empty}\fi

%%%%%%%%% ABSTRACT
\begin{abstract}
The goal of text-to-image synthesis is to generate a visually realistic image that matches a given text description. In
practice, the captions annotated by humans for the same image have large variance in terms of contents and the choice
of words. The linguistic discrepancy between the captions
of the identical image leads to the synthetic images deviating from the ground truth. To address this issue, we propose a contrastive learning approach to improve the quality
and enhance the semantic consistency of synthetic images.
In the pretraining stage, we utilize the contrastive learning approach to learn the consistent textual representations
for the captions corresponding to the same image. Furthermore, in the following stage of GAN training, we employ the
contrastive learning method to enhance the consistency between the generated images from the captions related to the
same image. We evaluate our approach over two popular
text-to-image synthesis models, AttnGAN and DM-GAN, on
datasets CUB and COCO, respectively. Experimental results have shown that our approach can effectively improve
the quality of synthetic images in terms of three metrics:
IS, FID and R-precision. Especially, on the challenging
COCO dataset, our approach boosts the FID significantly
by 29.60\% over AttnGAN and by 21.96\% over DM-GAN.
\end{abstract}

%%%%%%%%% BODY TEXT
\section{Introduction}
The objective of the text-to-image synthesis problem is to generate high-quality images from the specific text descriptions. It is a fundamental problem with a wide range of
practical applications, including art generation, image editing, and computer-aided design. Most recently proposed text-to-image synthesis methods~\cite{ zhang2017stackgan, zhang2018stackgan++, xu2018attngan,hong2018inferring,zhang2018photographic, qiao2019mirrorgan,zhu2019dm,yin2019semantics,li2019object,qiao2019learn,NEURIPS2019_1d72310e,cha2019adversarial,hinz2019generating,el2019tell,liang2020cpgan,tao2020df,cheng2020rifegan,tan2019text2scene, 
johnson2018image,
tseng2020retrievegan,
zhang2021cross}
are based on Generative Adversarial Networks (GANs)~\cite{goodfellow2014generative}. Conditioned on the text descriptions, the GAN-based models can generate realistic images with consistent semantic meaning. In practice,
one image is associated to multiple captions in the datasets. These text descriptions annotated by humans for the same image are highly subjective and diverse in terms of contents and choice of words. Additionally, some text descriptions do not even provide sufficient semantic information to guide the image generation. The linguistic variance and inadequacy between the captions of the identical image leads to the synthetic images conditioned on them deviating from
the ground truth.

To address this issue, we propose a novel contrastive
learning approach to improve the quality and enhance the
semantic consistency of synthetic images. In the image-text
matching task, we pretrain an image encoder and a text encoder to learn the semantically consistent visual and textual
representations of the image-text pair. Meanwhile, we learn
the consistent textual representations by pushing together
the captions of the same image and pushing way the captions of different images via the contrastive loss. The pretrained image encoder and text encoder are leveraged to extract consistent visual and textual features in the following
stage of GAN training. Then we also utilize the contrastive
loss to minimize the distance of the fake images generated
from text descriptions related to the same ground truth image while maximizing those related to different ground truth
images. We generalize the existing text-to-image models
to a unified framework so that our approach can be integrated into them to improve their performance. We evaluate our approach over two popular base models, AttnGAN~\cite{xu2018attngan} and DM-GAN~\cite{zhu2019dm} on datasets CUB~\cite{wah2011caltech} and COCO~\cite{lin2014microsoft}. The experimental results have shown that our approach can effectively improve the quality of the synthetic images in terms of Inception Score (IS) ~\cite{salimans2016improved}, Fr\'{e}chet Inception Distance (FID)~\cite{heusel2017gans}, and R-precisions~\cite{ xu2018attngan}.

The contributions of our work can be summarized as follows:
1) We propose a novel contrastive learning approach to learn the semantically consistent visual and textual representations in the image-text matching task. 2)
We propose a novel contrastive learning approach to enhance the semantic consistency of the synthetic images in the stage of GAN training. 3) Our approach can be incorporated into the existing
text-to-image models to improve their performance. Extensive experimental results demonstrate the effectiveness
of our approach. Our source code is publicly available  at \url{ https://github.com/huiyegit/T2I_CL}.

\begin{figure*}
\begin{center}
% \fbox{\rule{0pt}{2in} \rule{.9\linewidth}{opt}}
\includegraphics[width=0.9\textwidth]{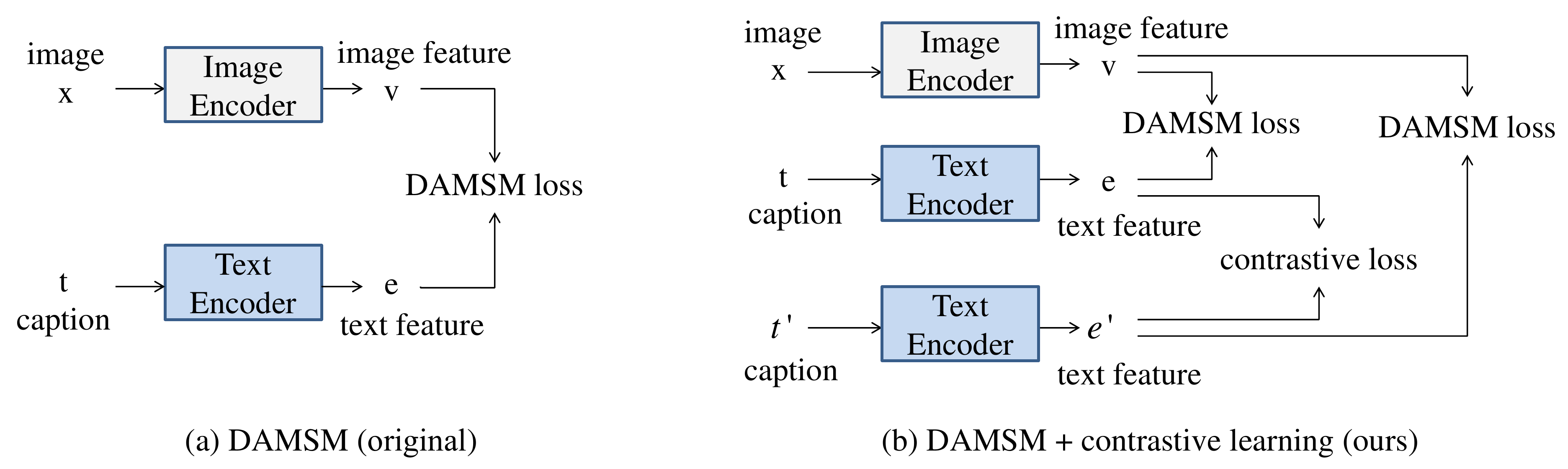}
\end{center}
\vskip -0.15in
\caption{Architectures of original DAMSM and our approach for image-text matching.}
\vskip -0.10in
\label{fig:text-contrastive}
\end{figure*}

\section{Related Work}
\subsection{Text-to-Image Generation}

Recently, a great number of studies~\cite{reed2016generative, zhang2017stackgan, zhang2018stackgan++, xu2018attngan,hong2018inferring,zhang2018photographic, qiao2019mirrorgan,zhu2019dm,yin2019semantics,li2019object,qiao2019learn,NEURIPS2019_1d72310e,cha2019adversarial,hinz2019generating,el2019tell,liang2020cpgan,tao2020df,cheng2020rifegan,ramesh2021zero, tan2019text2scene, 
johnson2018image,
tseng2020retrievegan,
zhang2021cross} present
promising results on the text-to-image synthesis task, most of which make use of GANs as the backbone model.
We briefly summarizes some of them that are most related to our approach. 
Zhang et al.~\cite{zhang2017stackgan} propose the stacked GAN architecture which produces images from low-resolution to high-resolution. AttnGAN~\cite{xu2018attngan} presents an attention mechanism, where the Deep Attentional Multimodal Similarity Model (DAMSM) is able to compute the similarity between the generated image and the caption using both the global sentence level information and the fine-grained word level information. DM-GAN~\cite{zhu2019dm} introduces a dynamic memory generative adversarial network to generate high-quality images. It utilizes a dynamic memory module to refine the initial generated image, a memory writing gate to highlight the relevant text information and a response gate to update image representations. SD-GAN~\cite{yin2019semantics} employs a Siamese structure with a pair of texts as input and trains the model with the contrastive loss. The conditional batch normalization is adopted for fine-grained image generation. Compared with the Siamese structure in SD-GAN, our approach is derived from the recent development of the contrastive learning paradigm, therefore, it has the advantage of better performance and less computational cost.  Furthermore, we generalize our approach so that it can be applied to the existing GAN-based models for text-to-image synthesis. XMC-GAN ~\cite{zhang2021cross} has also applied the contrastive
learning approach in the text-to-image generation. However, the objectives of contrastive loss in our approach are
different from those of XMC-GAN. We compute the contrastive losses of the caption-caption pair and the fake-fake
pair, which are complementary to the contrastive losses in
XMC-GAN.   In this work, we choose AttnGAN and DM-GAN as the base models to evaluate our approach.

\subsection{Image-Text Matching}

The text-to-image synthesis involves the subtask image-text matching, which refers to learning the joint image-text
representation to maximize the semantic similarity for an
image-sentence pair.  Liwei et al.~\cite{wang2016learning} and Michael  et al.~\cite{wray2019fine} have leveraged the triplet loss to learn the joint image-text embedding for the image-text retrieval and the video-text  representation for the video-text action retrieval task, respectively.  Tao et al.~\cite{xu2018attngan} propose the Deep Attentional Multimodal Similarity Model (DAMSM) to learn the
fine-grained image-text representation for text-to-image synthesis. DAMSM (Figure~\ref{fig:text-contrastive}a) trains an image encoder and a text encoder jointly to encode sub-regions of the image and words of the sentence to a common semantic space, and computes a fine-grained image-text matching loss for image generation. However, the variations exist
in the text representations corresponding to the same image, which leads to the generated images deviating from the
ground truth image. To address this issue, we utilize the
contrastive learning approach to push together the text representations related to the identical image and push away
the text representations related to different images.

\subsection{Contrastive Learning}

Contrastive learning has recently attracted great interest due to its empirical success in self-supervised representation learning in computer vision. In the last two years,  
various contrastive methods~\cite{saunshi2019theoretical,chen2020simple,he2020momentum,chen2020improved,khosla2020supervised,tian2020makes,robinson2020contrastive,chuang2020debiased,hassani2020contrastive,henaff2020data,kalantidis2020hard,misra2020self,tian2019contrastive,wang2020understanding} for visual representations have been proposed. SimCLR~\cite{chen2020simple} presents three major findings to learn better representations, including composition of data augmentations, a learnable nonlinear transformation between the representation and the contrastive loss, and large batch size and training step. 
Similar to SimCLR, we adopt the simple contrastive learning framework. 
In order to integrate the contrative learning approach into the GAN-based models with simple implementation and small computational cost, our approach does not have the learnable nonlinear transformation or large batch size. We set the same training batch sizes as our baselines.

\section{Method}
 
\subsection{Contrastive Learning for Pre-training}
\label{cl_text}

In the text-to-image synthesis task, the purpose of image-text matching is to learn the text representations which are semantically consistent with the corresponding images.  Since the text representations will be leveraged as the conditions to guide the image generation, it is beneficial to develop a more effective pre-training approach to improve the quality of synthetic images. 

Inspired by recent contrastive learning algorithms, we propose a novel approach for the pre-training of image-text matching. We learn the textual representations to match the visual representations via the DAMSM loss. Moreover, we train the textual representations by pushing together the captions corresponding to the same image and pushing away the captions corresponding to different images via the contrastive loss. As illustrated in Figure~\ref{fig:text-contrastive}b, our framework consists of the following three major components.

\textbf{Data sampling.} At each training step, we sample a minibatch of images $\bs{x}$, captions $\bs{t}$ and captions $\bs{t}^\prime $, where both captions $\bs{t}$ and $\bs{t}^\prime$ are corresponding to images $\bs{x}$. For image-text matching, we consider two positive image-caption pairs $(x_i, t_i )$ and $(x_i, t_i^\prime)$ for each image $x_i$ to calculate the DAMSM loss.  Furthermore, we consider the caption-caption pair $(t_i, t_i^\prime )$ as the positive pair to calculate the contrastive loss.

\textbf{Image encoder $f$ and text encoder $g$.} We adopt an image encoder $ f$ to extract the visual vector representations and sub-region features from the image samples. Furthermore, we utilize a text encoder $g$ to extract the textual vector representations and word features from the text samples. The  text encoder $g$ is shared in the framework. Our  architecture has the flexibility of allowing various choices of deep neural network models.
 To have a fair comparison with the baselines, we adopt the same Inception-v3~\cite{szegedy2016rethinking} and Bi-directional Long Short-Term
Memory (Bi-LSTM)~\cite{schuster1997bidirectional} to instantiate the image encoder $f$ and text encoder $g$.

\textbf{Loss function.} Similar to the baselines, we adopt the DAMSM loss as image-text matching loss. 
Moreover, we define the contrastive loss on pairs of two branches of input captions.  We compute the contrastive loss to minimize the distance of textual representations  related to the same image while maximizing those related to different images. We utilize the Normalized Temperature-scaled Cross Entropy Loss (NT-Xent)~\cite{sohn2016improved,wu2018unsupervised,chen2020simple} as the contrastive loss.
 Given a pair, let $sim(a, b) = a^T b / \|a \| \|b\|$ denote the dot product between $l_2$ normalized $a$ and $b$. Then the loss function for the $i$th sample is defined as
\begin{equation}
\label{eqn:equation_1}
 L(i)=-\log\frac{\exp(sim(u_i, u_j)/\tau)}  {\sum_{k=1}^{2N}\mathbbm{1}_{k\neq i}\exp(sim(u_i, u_k)/\tau)},
\end{equation}
where the $i$th and $j$th sample make the positive pair, $\mathbbm{1}_{k\neq i}$ is an indicator function whose value is 1 iff ${k\neq i}$ , $\tau$ denotes a temperature parameter and $N$ is the batch size (e.g., $N$ images and $2N$ captions). The overall contrastive loss is computed across all positive pairs in a minibatch, which can be defined as
\begin{equation}
\label{eqn:equation_2}
 L_c= \frac{1}{2N} \sum_{i=1}^{2N}  L(i)
\end{equation}
% \vspace*{-.2cm}
Algorithm~\ref{alg:Alg.1} summarizes the proposed method.
% to pre-train the image-text matching task.

% \begin{algorithm}[H] 
\begin{algorithm}[t]  
\small
\caption{Contrastive learning for image-text matching} \label{alg:Alg.1}  
\begin{algorithmic}[1]
\REQUIRE{Batch size $N$, temperature $\tau$, image encoder $f$, text encoder $g$}
\ENSURE{Optimized image encoder $f$ and text encoder $g$}
\STATE 
{\bf for} \{1, $\cdots$, \#  of  training iterations\} do \\
\STATE 
\ \ \ Sample a minibatch of images  $\bs{x}$
\STATE 
\ \ \ Sample a minibatch  of captions $\bs{t}$ associated with $\bs{x}$
\STATE 
\ \ \ Sample another minibatch of captions $\bs{t}^\prime$ associated with $\bs{x}$

\STATE 
\ \ \ $\bs{v} = f(\bs{x})$  \COMMENT{image representation}
\STATE 
\ \ \ $\bs{e} = g(\bs{t})$  \COMMENT{text representation}
\STATE 
\ \ \ $\bs{e}^\prime = g(\bs{t}^\prime)$ \COMMENT{text representation}
\STATE 
\ \ \ $\mathcal{L}_1 = \text{DAMSM}(\bs{v},\bs{e}) $ \COMMENT{image-text matching loss}  
\STATE 
\ \ \ $\mathcal{L}_2 = \text{DAMSM}(\bs{v},\bs{e}^\prime) $ 
\COMMENT{image-text matching loss} 

\STATE 
\ \ \ $\mathcal{L}_c$ = \text{NT-Xent}$(\bs{e},\bs{e}^\prime)$ \COMMENT{Equation~\ref{eqn:equation_2}} 

\STATE 
\ \ \ $\mathcal{L} = \mathcal{L}_1 + \mathcal{L}_2 + \mathcal{L}_c $

\STATE 
\ \ \ Update networks $f$ and $g$ to minimize $\mathcal{L}$
\STATE 
{\bf end for}\\
% \UNTIL{n=N} 

\end{algorithmic}  
\end{algorithm}
% \vspace*{-.4cm}

\subsection{Contrastive learning for GAN training}

In practice, the captions annotated by humans for the same image have large variance in terms of contents and the choice of words, especially when the scenes are complex. The linguistic discrepancy between the captions of the identical image leads to the synthetic images conditioned on them deviating from the ground truth.  Inspired by recent contrastive learning approaches, we apply the contrastive learning method to enhance the consistency between the generated images from the captions related to the same image and motivate them to be closer to the ground truth. Our approach consists of the following three major components.

\begin{figure*}
\begin{center}
% \fbox{\rule{0pt}{2in} \rule{.9\linewidth}{opt}}
\includegraphics[width=0.92\textwidth]{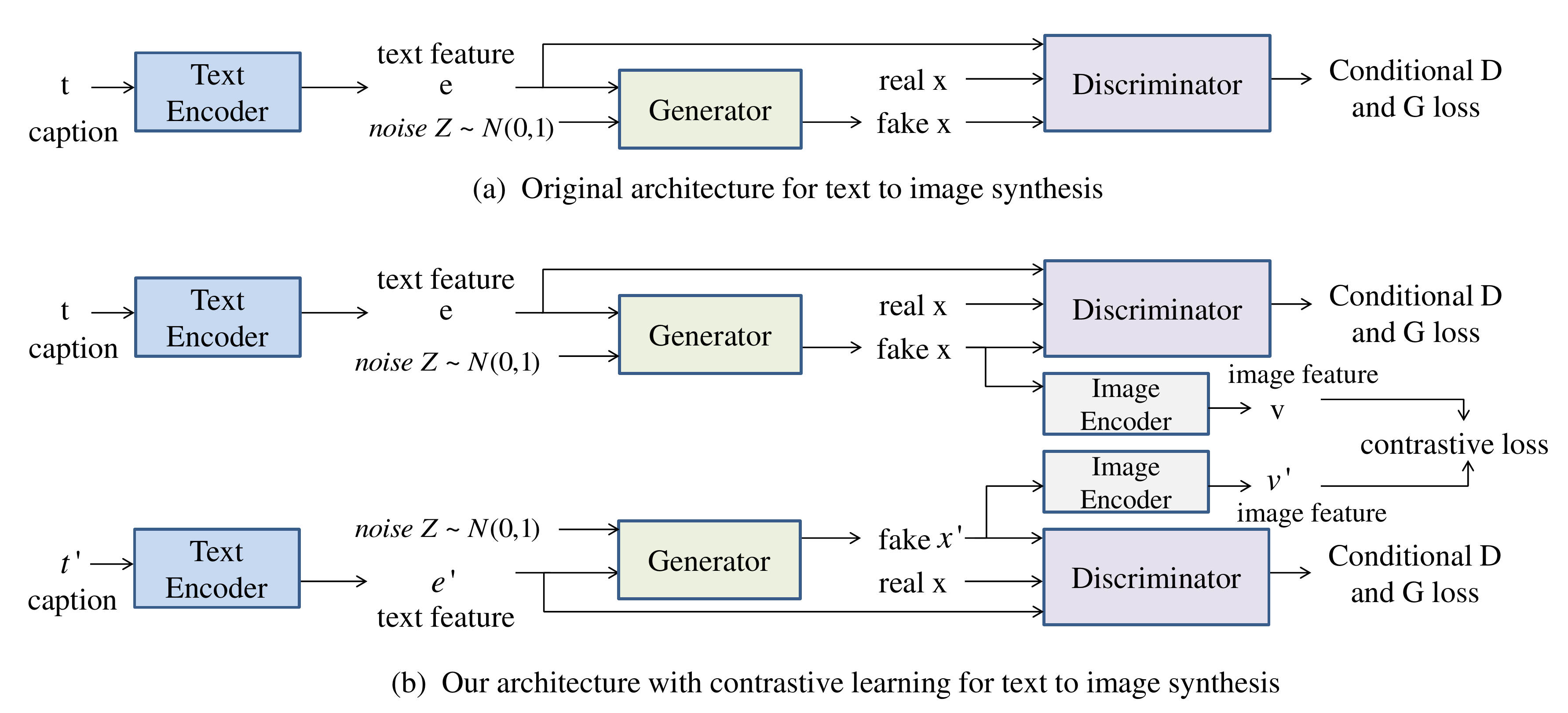}
\end{center}
\vskip -0.10in
\caption{Architectures of original approach and our approach for text to image synthesis.}
\label{fig:gan_contrastive}
\vskip -0.05in
\end{figure*}

\textbf{Data sampling.} The data sampling approach is similar to the one in the pre-training stage. At each training step, we sample a minibatch of images $\bs{x}$, captions $\bs{t}$ and captions $\bs{t}^\prime$ in the same way as in Section~\ref{cl_text}. The deep generative model outputs the fake images $\bs{x}$ and $\bs{x}^\prime$ conditioned on captions $\bs{t}$ and $\bs{t}^\prime$, respectively. Then we consider the image-image pair $(x_i, x_i^\prime)$ as the positive pair in the contrastive learning.

\textbf{Model architecture.} In this section, we derive our framework from the vanilla GANs step by step. GANs are a family of powerful generative models that estimate the data  distribution through an adversarial learning process, in which a generator network $G$ produces synthetic data given the input noise $z$ and a discriminator network $D$ distinguishes the true data from the generated data. The generator $G$ is optimized to output realistic samples to fool the discriminator $D$. Formally, the generator $G$ and discriminator $D$ are following the minimax objective: 
\begin{equation}
\label{eqn:gan}
\min_{G} \max_{D} \underset{x \sim P_r}{\mathbb{E}} [\log(D(x))] +\underset{z \sim P_z}{\mathbb{E}} [\log(1-D(G(z)))],
\end{equation}
where $x$ is a real sample from the data distribution $P_r$, and the input $z$ is sampled from some prior distribution $P_z$, such as a uniform or Gaussian distribution.

 Most of the recent works~\cite{ zhang2017stackgan, zhang2018stackgan++, xu2018attngan,hong2018inferring, qiao2019mirrorgan,zhu2019dm,yin2019semantics,li2019object,qiao2019learn,NEURIPS2019_1d72310e,cha2019adversarial,hinz2019generating,el2019tell,liang2020cpgan,tao2020df,cheng2020rifegan} on the text-to-image synthesis problem are based on GANs. We generalize these approaches to a unified framework, as shown in Figure~\ref{fig:gan_contrastive}a. The framework is extended from the GANs with the auxiliary information $e$, which is encoded from the input caption $t$ by a pre-trained text encoder $g$.  Then both the generator $G$ and discriminator $D$ are conditioned on this textual condition $e$. The training process is similar to the standard GANs with the following objective:
\begin{equation}
\label{eqn:cond-gan}
\begin{adjustbox}{max width=215pt}
$
\underset{G}{\min}\:
\underset{D}{\max} \!\! \underset{\:\:x \sim P_r}{\mathbb{E}}\![\log(D(x, e))]\!+\!\!\!\underset{\:z \sim P_z}{\mathbb{E}} [\log(1\!-\!D(G(z,e),e))] 
 $
 \end{adjustbox}
\end{equation}

\begin{algorithm}[t]  
\small
\caption{Contrastive learning for GAN  training}
\label{alg:gan}  
\begin{algorithmic}[1]
\REQUIRE{Batch size $N$, temperature $\tau$, coefficient $\lambda_c$, generator $G$, discriminator $D$, pre-trained image encoder $f$, pre-trained text encoder $g$.}
\ENSURE{ Optimized $G$ and $D$.}
\STATE 
{\bf for} \{1, $\cdots$, \# of training iterations\} do \\
\STATE 
\ \ \ Sample a minibatch  of images  $\bs{x}\sim P_r$
\STATE 
\ \ \ Sample a minibatch  of latent variable $\bs{z}\sim P_z$ 

\STATE 
\ \ \ Sample a minibatch of captions $\bs{t}$ associated with $\bs{x}$

\STATE 
\ \ \ Sample another minibatch of captions $\bs{t}^\prime$  associated with $\bs{x}$

\STATE 
\ \ \ $\bs{e} = g(\bs{t})$

\STATE 
\ \ \ $\bs{e}^\prime = g(\bs{t}^\prime)$ 

\STATE 
\ \ \  $ \mathcal{L}_{D1} =  \frac{1}{N} \sum\nolimits_{i=1}^{N}  \lbrack\log D(x_i, e_i) + \log (1- D(G(z_i, e_i), e_i)) \rbrack  $  

\STATE 
\ \ \  $ \mathcal{L}_{D2} =  \frac{1}{N} \sum\nolimits_{i=1}^{N}  \lbrack\log D(x_i, e_i^\prime) + \log(1- D(G(z_i, e_i^\prime), e_i^\prime)) \rbrack$ 

\STATE 
\ \ \ $ \mathcal{L}_{D} = 
\mathcal{L}_{D1} + \mathcal{L}_{D2}$

\STATE
\ \ \ Update $D$ to minimize $\mathcal{L}_{D}$

\STATE 
\ \ \ Sample noise $\bs{z}$, captions $\bs{t}$ and $\bs{t}^\prime$ as step 3, 4 and 5

\STATE 
\ \ \ Compute  $\bs{e}$, $\bs{e}^\prime$ as step 6 and 7  

\STATE 
\ \ \  $ \mathcal{L}_{G1} =  \frac{1}{N} \sum\nolimits_{i=1}^{N}\log (1- D(G(z_i, e_i), e_i)) $  

\STATE 
\ \ \  $ \mathcal{L}_{G2} =  \frac{1}{N} \sum\nolimits_{i=1}^{N}\log (1- D(G(z_i, e_i^\prime), e_i^\prime))   $ 

\STATE 
\ \ \ $\bs{v} = f(G(\bs{z}, \bs{e}))$  \COMMENT{image representation}

\STATE 
\ \ \ $\bs{v}^\prime = f(G(\bs{z}, \bs{e}^\prime))$  \COMMENT{image representation}

\STATE 
\ \ \ $\mathcal{L}_c$ = \text{NT-Xent} $(\bs{v},\bs{v}^\prime)$ \COMMENT{Equation~\ref{eqn:equation_2}} 

\STATE 
\ \ \ $ \mathcal{L}_{G} = 
\mathcal{L}_{G1} + \mathcal{L}_{G2} + \lambda_c  \mathcal{L}_c$ 

\STATE 
\ \ \ Update $G$ to minimize $ \mathcal{L}_{G}$
\STATE 
{\bf end for}\\
% \UNTIL{n=N} 
\end{algorithmic}  
\end{algorithm}

We further extend the generalized text-to-image framework to a Siamese structure and integrate the contrastive learning approach into it. As shown in Figure~\ref{fig:gan_contrastive}b, the image encoder $f$ takes  the fake images $x$ and $ x^\prime$ as input, and extracts the visual representations $v$ and $ v^\prime$ to compute the contrastive loss. The two branches of the architecture share the identical generator $G$, discriminator $D$, image encoder $f$ and text encoder $g$. The image encoder $f$ and text encoder $g$ are pre-trained in the image-text matching task and work in the evaluation mode in the phase of GAN training. 

\textbf{Loss function.}
In addition to the adversarial losses from Equation~\ref{eqn:cond-gan}, we define the contrastive loss on pairs of fake images generated from two branches of input captions. We utilize the contrastive loss to minimize the distance of the fake images generated from two text descriptions related to  the same image while maximizing those related to different images. We apply the same NT-Xent loss in Section~\ref{cl_text}.
Algorithm~\ref{alg:gan} summarizes the proposed method.
% for the GAN training.

\section{Experiments}
\noindent
\textbf{Datasets.} 
Following the previous works, we evaluate our approach on datasets CUB~\cite{wah2011caltech}
and COCO~\cite{lin2014microsoft}.
 The CUB dataset contains 200 bird species with 11,788 images, where 150 species with 8,855 images are used as the training data, and the remaining 50 species with 2,933 images as the test data. Each image has 10 related captions in the CUB dataset. The COCO dataset has 80k images for training  and 40k images for evaluation. Each image has 5 related captions in the COCO dataset.

\noindent
\textbf{Evaluation Metric.} We choose the  
Inception Score (IS) ~\cite{salimans2016improved}, Fr\'{e}chet Inception Distance (FID)~\cite{heusel2017gans}, and R-precisions~\cite{ xu2018attngan} as the quantitive metrics to evaluate the performance. 
The IS calculates the KL-divergence between the conditional and marginal probability distributions.
In general, a larger IS indicates the generative model can synthesize fake images with better diversity and quality.
 The FID computes the Fr\'{e}chet distance between synthetic and real images in the feature space  extracted from the pre-trained Inception v3 model. A smaller FID indicates the  synthetic data is more realistic and similar to the true data. R-precision calculates the precision of the image-text retrieval task to evaluate to what extent the synthetic images match the input captions. The higher R-precision indicates the generated images have greater consistency with the text descriptions. After training, the model generates 30,000 images conditioned on the captions in the test set for evaluation. The source code to calculate the three metrics is from the public website\footnote{\url{https://github.com/MinfengZhu/DM-GAN}}.
 
 Note that several previous works~\cite{ li2019object,tao2020df} have pointed out that IS can not provide useful guidance to evaluate the quality of the synthetic images on dataset COCO. Nevertheless, we still report the IS on COCO as the auxiliary metric. We consider the FID as the primary metric among the three in terms of robustness and effectiveness.

\noindent
\textbf{Implementation details.}
We choose two popular text-to-image synthesis models, AttnGAN~\cite{xu2018attngan} and DM-GAN~\cite{zhu2019dm}, to evaluate our approach.
Note that both AttnGAN and DM-GAN are the stacked architecture with 3 generator-discriminator pairs. To reduce the computational cost, we only calculate the contrastive loss of the   fake images of size 256x256 from the last generator.

We retain the setting of parameters in the original baselines except the $\lambda$ value used in AttnGAN. We find that $\lambda=5$ is reported in the paper and used in the source code. However, when we ran the source code of AttnGAN with this value, the R-precision is about $58.80$, which has a large difference from $67.21$ reported in the paper. When we changed $\lambda$ to 10, we got $67.00$ for R-precision, which is consistent with the one reported in the paper as well as the IS score and FID. We believe there is a typo of $\lambda$ value in the AttnGAN paper. Therefore, we set $\lambda=10$ and adopt it in all of our experiments.

Following the same setting of the configuration files of the baselines, we train our novel model based on AttnGAN with 600 epochs on CUB and 120 epochs on COCO, and the other model based on DM-GAN with 800 epochs on CUB and 200 epoch on COCO. We evaluate the IS, FID and R-precision of the checkpoint every 50 epochs on CUB and 10 epochs on COCO. We choose the checkpoint with the best FID and report the corresponding IS and R-precision.

\begin{table}[t]
% \vskip -0.25in
\begin{center}
% \vskip +0.1in
\begin{adjustbox}{width=0.47\textwidth}
\begin{tabular}{ l c c  c c  c  }
\toprule
Method & Dataset & IS $\uparrow$ & FID $\downarrow$ &  R-Precision $\uparrow$ \\
\midrule
AttnGAN$^\ast$ & CUB & 4.33 $\pm$ .07 & 20.85 & 67.09 $\pm$ .83   \\
AttnGAN + CL & CUB & \textbf{4.42 $\pm$ .05} & \textbf{16.34} & \textbf{69.64 $\pm$ .63}   \\
\hline
AttnGAN$^\ast$ & COCO & 23.71 $\pm$ .38 & 33.99 & 83.97 $\pm$ .78   \\
AttnGAN + CL& COCO & \textbf{25.70 $\pm$ .62
} & \textbf{23.93
} & \textbf{86.55 $\pm$ .51
}   \\
\bottomrule
\end{tabular}
\end{adjustbox}

\end{center}
\vskip -0.05in
\caption{Comparison of our approach and AttnGAN over the datasets CUB and COCO. $\uparrow$ denotes the higher value the better quality. $\downarrow$ denotes the lower value the better quality.  $^\ast$ indicates the results are obtained from the pre-trained model released publicly by the authors. The bold font represents better performance. CL denotes the proposed constrastive learning approach in this work.}\label{table:attngan_result}
% \vskip -0.2in
\end{table}

\begin{table}[t]
% \vskip -0.25in
\begin{center}
% \vskip +0.1in
\begin{adjustbox}{width=0.47\textwidth}
\begin{tabular}{l c c c c  c }
\toprule
Method & Dataset & IS $\uparrow$ & FID $\downarrow$ &  R-Precision $\uparrow$ \\
\midrule
DM-GAN$^\ast$ & CUB & 4.66 $\pm$ .06
 & 15.10 & 75.86 $\pm$ .83   \\
DM-GAN + CL & CUB & \textbf{4.77 $\pm$ .05} & \textbf{14.38} & \textbf{78.99 $\pm$ .66}   \\
\hline
DM-GAN$^\ast$ & COCO & 32.37$\pm$ .29 & 26.64 & 92.09 $\pm$ .50   \\
DM-GAN + CL& COCO & \textbf{33.34 $\pm$ .51 
} & \textbf{20.79
} & \textbf{93.40 $\pm$ .39
}   \\
\bottomrule
\end{tabular}
\end{adjustbox}

\end{center}
\vskip -0.05in
\caption{Comparison of our approach and DM-GAN over the datasets CUB and COCO. The notations $\uparrow$, $\downarrow$, $^\ast$, bold font and CL have the same meanings as the ones in Table~\ref{table:attngan_result}}
\label{table:dmgan_result}
\vskip -0.1in
\end{table}

\begin{figure*}
\begin{center}
% \fbox{\rule{0pt}{2in} \rule{.9\linewidth}{opt}}
\includegraphics[width=1.0\textwidth]{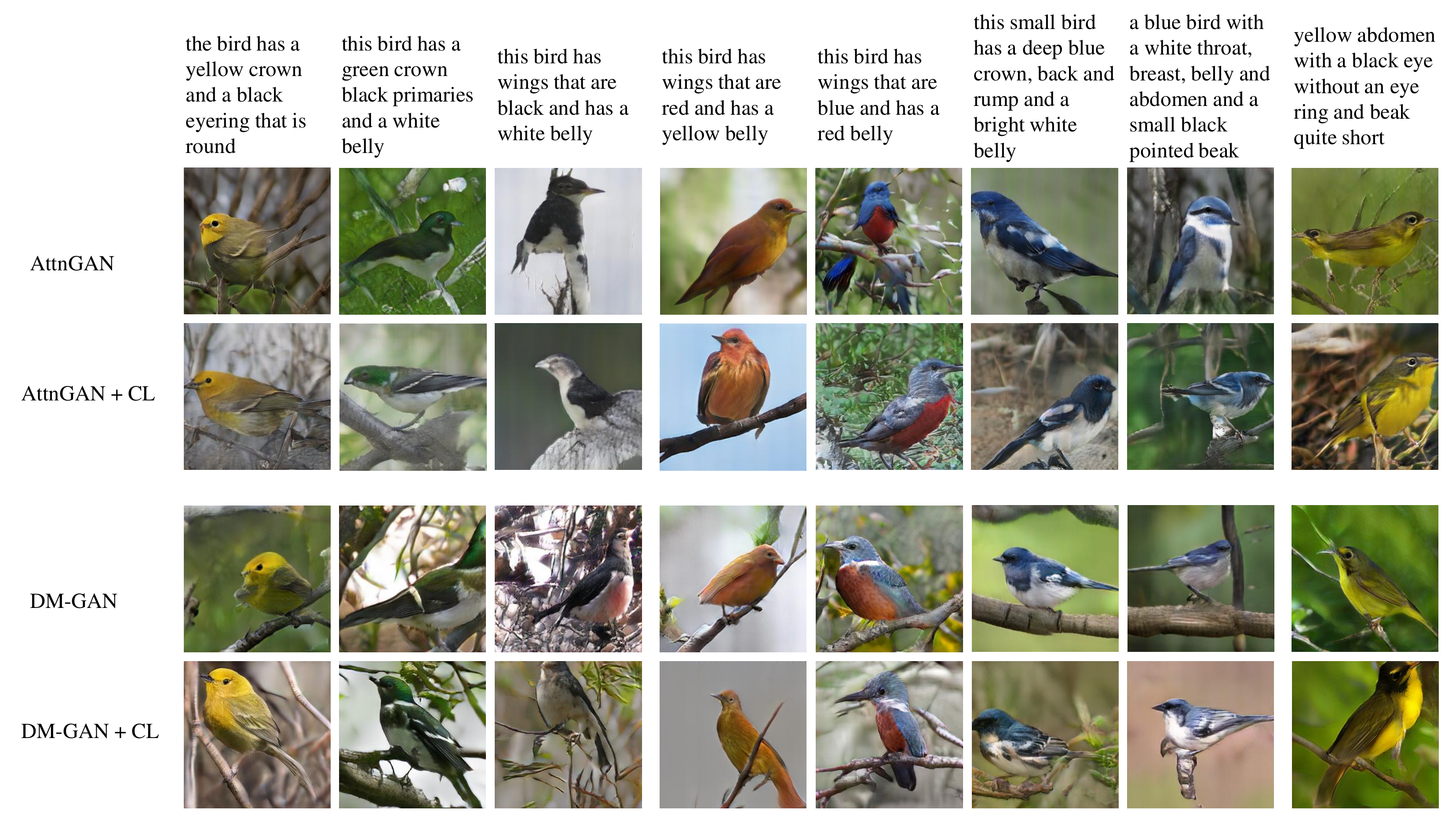}
\end{center}
\vskip -0.1in
\caption{Comparison of example images between our approach and baselines on the CUB dataset.}

\vskip 0.1in

\label{fig:image1}
\end{figure*}

\begin{figure*}
\begin{center}
% \fbox{\rule{0pt}{2in} \rule{.9\linewidth}{opt}}
\includegraphics[width=1.0\textwidth]{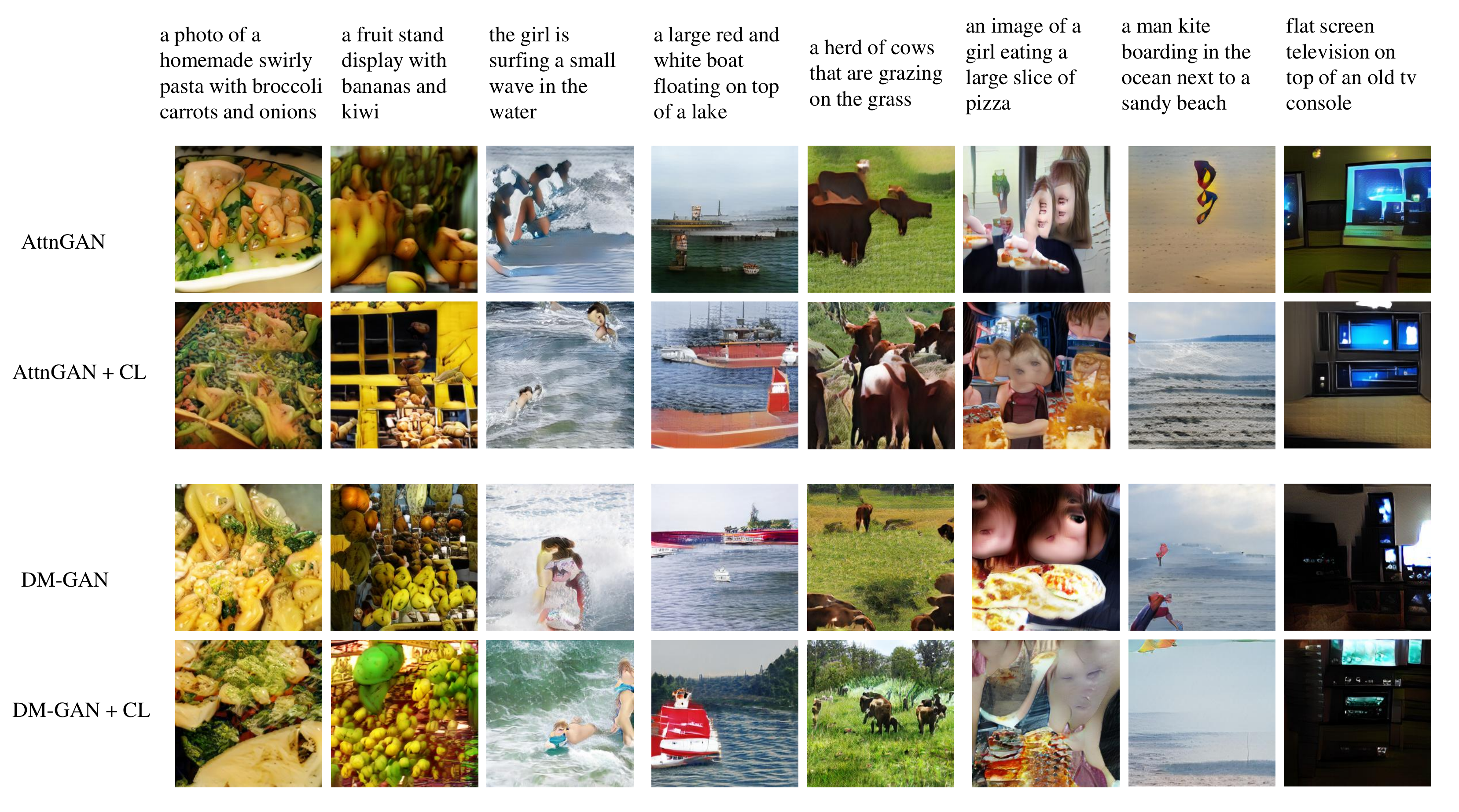}
\end{center}
\vskip -0.15in
\caption{Comparison of example images between our approach and baselines on the COCO dataset.}
\label{fig:image2}
\end{figure*}

\begin{figure*}
\begin{center}
% \fbox{\rule{0pt}{2in} \rule{.9\linewidth}{opt}}
\includegraphics[width=1.0\textwidth]{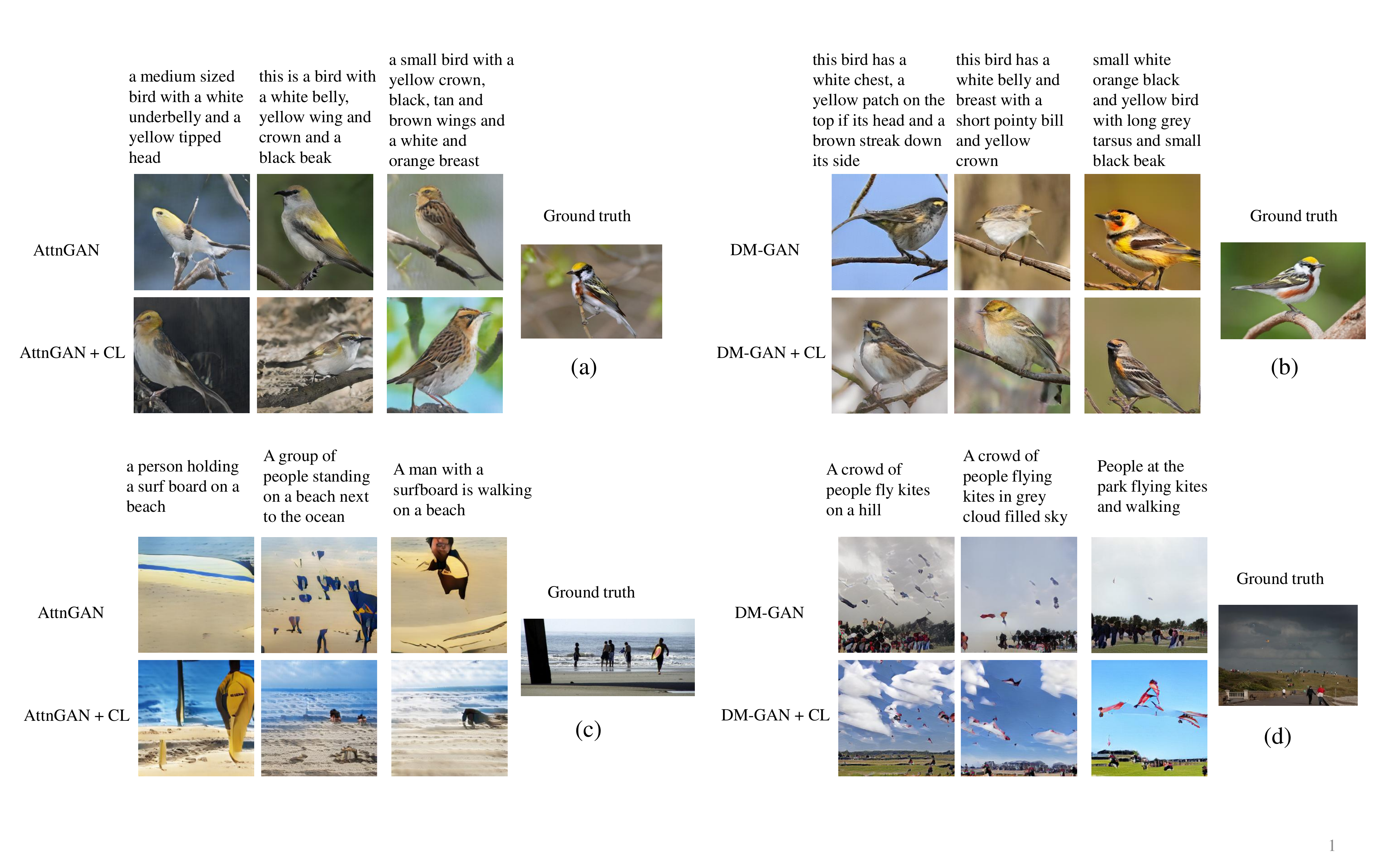}
\end{center}
\vskip -0.55in
\caption{Comparison of example images between our approach and baselines.}
\label{fig:image3}
\vskip -0.15in
\end{figure*}

\subsection{Text-to-Image Quality}

We apply our contrastive learning approach to two  baselines AttnGAN~\cite{xu2018attngan} and DM-GAN~\cite{zhu2019dm}, and compare the performances with them over datasets CUB and COCO. The experimental results are reported in Table~\ref{table:attngan_result} and~\ref{table:dmgan_result}. 

As shown in Table~\ref{table:attngan_result}, our approach improves the three metrics IS, FID and R-precision over the datasets CUB and COCO. The IS is improved from 4.33 to 4.42 on CUB and from 23.71 to 25.70 on COCO. For the relatively more suitable metric FID, our approach boosts the baseline AttnGAN significantly by 21\% on CUB and 29.60\% on COCO, respectively. Meanwhile, our approach achieves higher R-precision with the gain of 2.55 on CUB and 2.58 on COCO. As shown in Table~\ref{table:dmgan_result}, in comparison to DM-GAN, our approach also improves the three metrics IS, FID and R-precision over CUB and COCO. The IS is improved from 4.66 to 4.77 on CUB and from 32.37 to 33.34 on COCO. Regarding the metric FID , our approach has the value of 14.38 with a small gain of 0.72 on CUB, while boosts the baseline DM-GAN significantly by 21.96\% on COCO. Furthermore, our approach achieves better R-Precision with the improvement of 3.13 on CUB and 1.31 on COCO.

 The dataset COCO is more challenging than CUB, as it has more complex scenes and the captions have greater variance to describe the identical image. However, it is noteworthy that our approach significantly improves the FID by 29.60\% over the baseline AttnGAN and 21.96\% over DM-GAN. In summary, the quantitative experimental results demonstrate that our contrastive learning approach can effectively improve the quality and enhance the consistency of the synthetic images generated from diverse captions.

\subsection{Visual Quality}
To further compare our proposed approach with the baselines, we visualize the synthetic images generated from the typical example captions. As shown in Figure~\ref{fig:image1}, compared with the baseline AttnGAN, the images generated from our approach are more realistic and better match with the text descriptions in most cases. In the 8th column, the bird in the image from AttnGAN fails seriously with two heads, while the one from our approach has the reasonable appearance. In the 2nd column, we can see the vivid green crown in the bird from our approach, which matches the description ``green crown" well, while the image from AttnGAN does not show this feature of the bird.  As shown in Figure~\ref{fig:image1}, the comparison between our approach and the baseline DM-GAN is similar to previous comparison. In the 3rd column, the image from our approach has the correct white belly to match the text description ``while belly", while the image from DM-GAN has the additional incorrect red color in the belly. Figure~\ref{fig:image2} shows the example images on COCO from our approach and the baselines AttnGAN and DM-GAN. It is  challenging to generate photo-realistic images for the models showed in the figure. However, compared with the baselines, the images generated from our approach are more realistic and better match with the text descriptions in some cases. In the 4th column, the boat in the image from our approach has the red and while color, which aligns with the caption, while the image from AttnGAN does not show the red boat. In the 5th column, the image from our approach has better shape of cows than the one from AttnGAN. As shown in the 3rd column, the image from our approach has the basic shape of a girl, while it can not be observed in the image from DM-GAN at all. 

We also visualize the example images generated from multiple captions corresponding to the same ground truth image. Compared with the baselines AttnGAN and DM-GAN, the images generated from our approach are more realistic and closer to the ground truth images. As shown in the 1st column of Figure~\ref{fig:image3}(a), the image from our approach has the black color in the wings, which is consistent with the ground truth image. Although the 1st caption does not have the text description of \enquote{black color} explicitly, our model is still able to train its semantic text embedding to contain this information from other captions related to the same image via our contrastive learning approach. Another similar example is shown in Figure~\ref{fig:image3}(c). The 3rd caption does not provide the text description of \enquote{the ocean}, the image from our approach can still have the ocean scene to be consistent with the ground truth, while the baseline DM-GAN is not able to achieve this.

% \begin{figure*}
% \begin{center}
% % \fbox{\rule{0pt}{2in} \rule{.9\linewidth}{opt}}
% \includegraphics[width=0.9\textwidth]{LaTeX/figure/image4.pdf}
% \end{center}
% \caption{The COCO dataset}
% \label{fig:short}
% \end{figure*}

\begin{table}[t]
% \vskip -0.25in
\centering
% \vskip +0.1in
\begin{adjustbox}{width=0.47\textwidth}
\begin{tabular}{ l c c  c c  c  }
\toprule
Method & Dataset & IS $\uparrow$ & FID $\downarrow$ &  R-Precision $\uparrow$ \\
\midrule
AttnGAN$^\dagger$ & CUB & 4.29 $\pm$ .05 & 19.16 & 68.02  $\pm$ .98  \\

 + CL1 & CUB & 4.31 $\pm$ .02 & 17.97 & 69.11 $\pm$ .63  \\

+ CL1 + CL2 & CUB & \textbf{4.42 $\pm$ .05} & \textbf{16.34} & \textbf{69.64 $\pm$ .63}   \\
\hline

AttnGAN$^\dagger$ & COCO & 25.05 $\pm$ .64 & 30.67 & 84.24 $\pm$ .58  \\

 + CL1 & COCO & \textbf{25.87 $\pm$ .41} & 26.89 & 85.93  $\pm$ .63 \\

+ CL1 + CL2 & COCO & 25.70 $\pm$ .62
& \textbf{23.93
} & \textbf{86.55 $\pm$ .51
}   \\

\bottomrule
\end{tabular}
\end{adjustbox}

 \vskip5pt
\caption{Ablation study of our approach on AttnGAN over the datasets CUB and COCO. $^\dagger$ indicates we retrain the model with the same setting of hyperparameters. CL1 and CL2 denote the constrastive learning approach in the pre-training and GAN training, respectively.}
\label{table:attngan_ablation}
 \vskip 0.2in
% \vskip -0.25in
 
\vskip-11pt

\begin{adjustbox}{width=0.47\textwidth}
\begin{tabular}{l c c c c  c }
\toprule
Method & Dataset & IS $\uparrow$ & FID $\downarrow$ &  R-Precision $\uparrow$ \\
\midrule
DM-GAN$^\dagger$ & CUB & 4.67 $\pm$ .06
 & 15.55 & 75.88  $\pm$ .89 \\
 
 + CL1 & CUB & 4.71 $\pm$ .05
 & 14.56 & 76.74 $\pm$ .88   \\ 
 
+ CL1 + CL2 & CUB & \textbf{4.77 $\pm$ .05} & \textbf{14.38} & \textbf{78.99 $\pm$ .66}   \\
\hline

DM-GAN$^\dagger$ & COCO & 31.53 $\pm$ .39 & 27.04 & 91.82 $\pm$ .49  \\

 + CL1 & COCO & 30.98 $\pm$ .69 & 25.29 & 92.10  $\pm$ .56 \\

+ CL1 + CL2 & COCO & \textbf{33.34 $\pm$ .51
} & \textbf{20.79
} & \textbf{93.40 $\pm$ .39
}   \\

\bottomrule
\end{tabular}
\end{adjustbox}

  \vskip5pt
\caption{Ablation Study of our approach on DM-GAN over the CUB and COCO datasets. $^\dagger$, CL1 and CL2 have the same meanings as the ones in Table~\ref{table:attngan_ablation}.}
\label{table:dmgan_ablation}

\vskip -11pt
\end{table}

\subsection{Ablation Study}
In this work, the contrastive learning approach is applied in two stages: image-text matching and the training of GANs. We combine our novel approach for the image-text matching with the baselines AttnGAN and DM-GAN, and conduct experiments to evaluate the effectiveness of it. The experimental results are shown in Table~\ref{table:attngan_ablation} and ~\ref{table:dmgan_ablation}. Compared with the two baselines AttnGAN and DM-GAN, our contrastive learning approach for the image-text matching task can help improve the performance in terms of the IS, FID and R-precision on datasets CUB and COCO, with one exception that the IS of our approach is about 0.55 smaller than DM-GAN on COCO. When we add the contrastive learning approach for GAN training into our previous approach, our complete approach shows further improvements in terms of the IS, FID and R-precision on datasets CUB and COCO, with the exception that the IS of our complete approach based on AttnGAN is about 0.11 smaller than our previous one on COCO. The experimental results demonstrate that our contrastive learning approach in the image-text matching task and GAN training can help improve the performance of text-to-image synthesis, respectively. 

We adjust the hyperparameters to investigate the impact to the performance of our approach.  We  tune the weight $\lambda_c$ in \{0.1, 0.2, 0.5, 1.0, 2.0, 5.0, 10.0\} and the temperature $\tau$ in \{0.1, 0.2, 0.5, 1.0\} for the contrastive loss. The experiments are conducted on the CUB dataset. In each case, we evaluate the checkpoint every 50 epochs and choose the one with the best FID. Table~\ref{table:ablation_lambda} shows the results of weight $\lambda_c$. We find that FID is not very sensitive to this hyperparameter as well as IS and R-precision. When $\lambda_c$ is 0.2, the model has the best FID score 16.34, which is improved by 1.42, compared with the worst value 17.76.  Table~\ref{table:ablation_temperature} shows the results of temperature $\tau $. Similar to the weight $\lambda_c$, it can be observed that $\tau$ has a small impact to the performance of the model. The difference between the largest FID and smallest one is 1.34. 

\begin{table}[t]
% \vskip -0.25in
\begin{center}
% \vskip +0.1in

\begin{adjustbox}{width=0.47\textwidth}
\begin{tabular}{ l | c | c  | c | c  c  }
\toprule
Method & $\lambda_c $ & IS $\uparrow$ & FID $\downarrow$ &  R-Precision $\uparrow$ \\
\midrule
\multirow{7}{*}  {AttnGAN + CL} & 0.1 & 4.28 $\pm$ .05  & 17.60 & 70.06 $\pm$ .84  \\

 & 0.2 & 4.42 $\pm$ .05 & \textbf{16.34} & 69.64 $\pm$ .63   \\

 & 0.5 & 4.35 $\pm$ .05 & 17.76 & 69.08 $\pm$ .63   \\

 & 1.0 & 4.31 $\pm$ .07 & 16.72 & 69.28 $\pm$ .77  \\

 & 2.0 & 4.41 $\pm$ .05 & 16.90 & 68.79 $\pm$ .38   \\

 & 5.0 & 4.49 $\pm$ .08 & 17.19 & 67.64 $\pm$ .85   \\

 & 10.0 & 4.43 $\pm$ .07 & 17.53 & 70.25 $\pm$ .55   \\

\bottomrule
\end{tabular}
\end{adjustbox}

\end{center}
\vskip -0.10in
\caption{Ablation study on different choices of weight $\lambda_c $ for contrastive loss.}
\label{table:ablation_lambda}

\vskip -0.05in
\end{table}

\begin{table}[t]
% \vskip -0.25in
\begin{center}
% \vskip +0.1in

\begin{adjustbox}{width=0.47\textwidth}
\begin{tabular}{ l | c | c | c |  c   }
\toprule
Method & $\tau $ & IS $\uparrow$ & FID $\downarrow$ &  R-Precision $\uparrow$ \\
\midrule
  \multirow{4}{*}  {AttnGAN + CL}  & 0.1 & 4.44	$\pm$ .06 & 17.12 & 69.62 $\pm$ .78  \\
  & 0.2 & 4.44 $\pm$ .05 & 17.55 & 68.50 $\pm$ .85  \\

 & 0.5 & 4.42 $\pm$ .05 & \textbf{16.34} & 69.64 $\pm$ .63  \\

 & 1.0 & 4.43 $\pm$ .05 & 17.68 & 69.86 $\pm$ .92  \\

\bottomrule
\end{tabular}
\end{adjustbox}

\end{center}
\vskip -0.10in
\caption{ Ablation study on different choices of temperature $\tau $ for contrastive loss.}

\label{table:ablation_temperature}

\vskip -0.1in
\end{table}

%-------------------------------------------------------------------------

\section{Conclusion}
In this paper, we have shown how to incorporate the contrastive learning method into prior text-to-image models to improve their performance. Firstly, we train the image-text matching task to push together the textual representations corresponding to the same image through the contrastive loss. Furthermore, we employ the contrastive learning method to enhance the consistency between generated images from the captions related to the same image. We propose a generalized framework for the existing text-to-image models, and evaluate our approach on two baselines AttnGAN and DM-GAN. Extensive experiments demonstrate that our approach outperforms the two strong baselines in terms of three metrics. Especially, on the challenging COCO dataset, our approach boosts the FID significantly by 29.60\% over AttnGAN and by 21.96\% over DM-GAN. Since the image-text representation learning is a fundamental task, we believe our approach has potential applicability in a wide range of cross domain tasks, such as visual question answering, image-text retrieval as well as text-to-image synthesis. We leave the extension to these tasks as a future work.

\section*{Acknowledgements}

We would like to thank the anonymous reviewers for their comments and suggestions, which helped improve the quality of this paper. We would also gratefully acknowledge the support of VMware Inc. for its university research fund to this research.

{\small
\bibliographystyle{ieee_fullname}
\bibliography{main}
}

\end{document}